\documentclass[graybox]{svmult}

\usepackage{mathptmx}       
\usepackage{helvet}         
\usepackage{courier}        
\usepackage{makeidx}         
\usepackage{graphicx}        
\usepackage{multicol}        
\usepackage[bottom]{footmisc}

\usepackage{wrapfig}

\makeindex             

\begin{document}

\title*{Debbie, the Debate Bot of the Future}
\author{Geetanjali Rakshit, Kevin K. Bowden, Lena Reed, Amita Misra, Marilyn Walker}
\institute{
Geetanjali Rakshit, Kevin Bowden, Lena Reed, Amita Misra, Marilyn Walker
\at Natural Language and Dialogue Systems Lab, University of California, Santa Cruz, \email{\{grakshit, kkbowden, lireed, amisra2,  mawalker\}@ucsc.edu}}

\date{}

\maketitle
\abstract{
Chatbots are a rapidly expanding application of dialogue systems with companies switching to bot services for customer support, and new applications for users interested in casual conversation.  One style of casual conversation is argument; many people love nothing more than a good argument. Moreover, there are a number of existing corpora of argumentative dialogues, annotated for agreement and disagreement, stance, sarcasm and argument quality. This paper introduces Debbie, a novel arguing bot, that selects arguments from conversational corpora, and aims to use them appropriately in context. We present an initial working prototype of Debbie, with some preliminary evaluation and describe future work.
}

\section{Introduction}
A chatbot or a conversational agent is a computer program that can converse with humans, via speech or text, with the goal of conducting a natural conversation, hopefully indistinguishable from a real human-to-human interaction. Chatbots are gaining momentum as more companies are
switching to bot services for customer care, but there is also an opportunity for different types of casual conversation.

Conversational agents can be broadly classified into retrieval-based and generative models. Retrieval-based methods have a repository of predefined responses and a mechanism to pick an appropriate response based on user input. They, therefore, can't generate a completely new response. Such methods are commonly used for ``help system'' chatbots, that target a predefined set of FAQs and responses. Another strategy is to use rule-based expression matching, and templated response generation, as in ELIZA or JULIA \cite{weizenbaum:1966,foner1997entertaining,mauldin1994chatterbots}. Most existing chatbots are task-oriented and their evaluation is based on the successful accomplishment of the task. 

There are many websites dedicated to debating controversial topics, and data from them have been structured and labeled in previous work. For example, we have access to trained models for labeling these argumentative conversations with attributes such as agreement, disagreement, stance, sarcasm, factual vs. feeling arguments, argument quality and argument facets. This data provides us with a rich source of conversational data for mining argumentative responses. We build on previous work in our lab on disagreement detection, classifying stance, identifying high quality arguments, measuring the properties and the
persuasive effects of factual vs. emotional arguments, and clustering
arguments into their facets or frames related to a particular topic \cite{lukinargument,Abbottetal16,MisraWalker13, Walkeretal12b,swanson2015argument,Misraetal16,Orabyetal15}.
  
In this work, we present Debbie, a novel arguing bot, that uses retrieval from existing conversations in order to argue with users. Debbie's main aim is to keep the conversation going, by successfully producing arguments and counter-arguments that will keep the user talking about the topic. Our initial prototype of Debbie works with three topics: death penalty, gun control, gay marriage. This paper focuses on our basic investigations on the initial prototype. While we are aware of other retrieval based chatbot systems \cite{duplessis:2016, Nio2014, Banchs:2012:ICD:2390470.2390477, Ameixa2014}, Debbie is novel in that it is the first to deal with argument retrieval.

\section{Data}
\label{sec:data}
Social media conversations are a good source of argumentative data but many sentences either do not express an argument or cannot be understood out of context and hence cannot be used to build Debbie's response pool. Swanson et al.(2015) created a large corpus consisting of 109,074 posts on the topics gay marriage (GM, 22425 posts), gun control (GC, 38102 posts), death penalty (DP, 5283 posts) by combining the Internet Argument Corpus(IAC) \cite{Walkeretal12a}, with dialogues
from online debate forums \footnote{http://www.createdebate.com/} \cite{swanson2015argument}.  It includes topic annotations, response characterizations (4forums), and stance. They build an argument quality regressor to rate the { \bf argument quality (AQ)} using a continuous slider ranging from hard (0.0) to easy to interpret (1.0). The AQ score is intended to reflect how easily the speaker's argument can be understood from the sentence without any context. Easily understandable sentences are assumed to be prime candidates for Debbie's response pool. 
Misra et al.(2016) note that a threshold of predicted AQ \textgreater 0.55 maintained both diversity and quality in the arguments \cite{Misraetal16}. For example, the sentence \textit{The death penalty is also discriminatory in its application what i mean is that through out the world the death penalty is disproportionately used against disadvantaged people} was given a score of 0.98.

We started with the Argument Quality (AQ) regressor from \cite{swanson2015argument}, which predicts a quality score for each sentence. The stance for these argument segments is obtained from IAC \cite{Abbottetal16}. We keep only stance bearing statements from the above dataset.
 Misra et al. (2016) had improved upon the AQ predictor from \cite{swanson2015argument}, giving a much larger and diverse corpus \cite{Misraetal16}. Since generating a cohesive dialogue is a challenging task, we first evaluated our prototype with hand labeled 2000 argument quality sentence pairs for the topic of death penalty obtained from \cite{Misraetal16}. We tested our model for both appropriateness of responses and response times. Once we had a working system for death penalty, we added the best quality 250 arguments for gay marriage and gun control, each, from the corpus of \cite{Misraetal16} (This had 174405 arguments from gay marriage and 258763 for gun control).

\section{Methodology}
\label{sec:methodology}

Debbie has domain knowledge of three hot button topics - death penalty, gay marriage and gun control. We created a database of statements for and against each topic, which serves as the source for Debbie's views.

The user picks a topic from a pool of topics and specifies his/her stance (for or against). As the user provides an argument, Debbie uses a similarity algorithm to retrieve a ranked list of the most appropriate counter-arguments, i.e., arguments opposing the user's stance. To speed up this retrieval process, we pre-create clusters of the arguments present in our database (described in section \ref{sec:methodology}). The most appropriate counter-arguments are calculated based on a similarity score, which, in this case, was the UMBC STS score \cite{han:2013}. The similarity algorithm takes as input two sentences and returns a real-valued score, which we use directly as the similarity between two argument statements. It uses a lexical similarity feature that combines LSA (Latent Semantic Similarity) word similarity, and WordNet knowledge, and can be run from a web API provided by the authors \cite{han:2013}.

Debbie's responses are stored for the duration of the chat. From the ranked list, the most appropriate response (having the highest similarity score), that has not already been used in the chat, is selected. Since we only have high quality arguments in our database, we expect Debbie's responses to be good in terms of grammatical correctness, on topic, etc. Another way of looking at appropriateness of a response is basically how adequate a retort it is to the user's view. Debbie sustains the debate until the user explicitly ends the chat. While there is a limited number of unique counter-arguments which Debbie can utilize, it would take the user substantial time to exhaust all possible responses.\\

\noindent{\bf Generating Clusters}

To create the clusters, we first generated a distance matrix of similarity scores for each topic and stance. A similarity score falls between 0 and 1. Using agglomerative clustering from scikit-learn, we created 15 clusters. We then identified the head of a cluster; the argument within each cluster, that best represents all of the statements within that cluster. We calculated this by finding the average distance for each statement in a cluster to all the statements in the cluster, and chose the one with the minimum average as the head. Hierarchical agglomerative clustering has been previously used for argument clustering by \cite{boltuzic2015identifying}.\\\\

\noindent{\bf Using the Clusters}

To speed up the response times by clustering, we compare the user's input to the head of each cluster. We find the cluster whose head has the highest similarity score and calculate the similarity score of each response within that cluster in order to return the most similar response. 

\begin{wrapfigure}{l}{0.6\textwidth}
\begin{center}
\vspace{-25pt}
\includegraphics[scale=0.21]{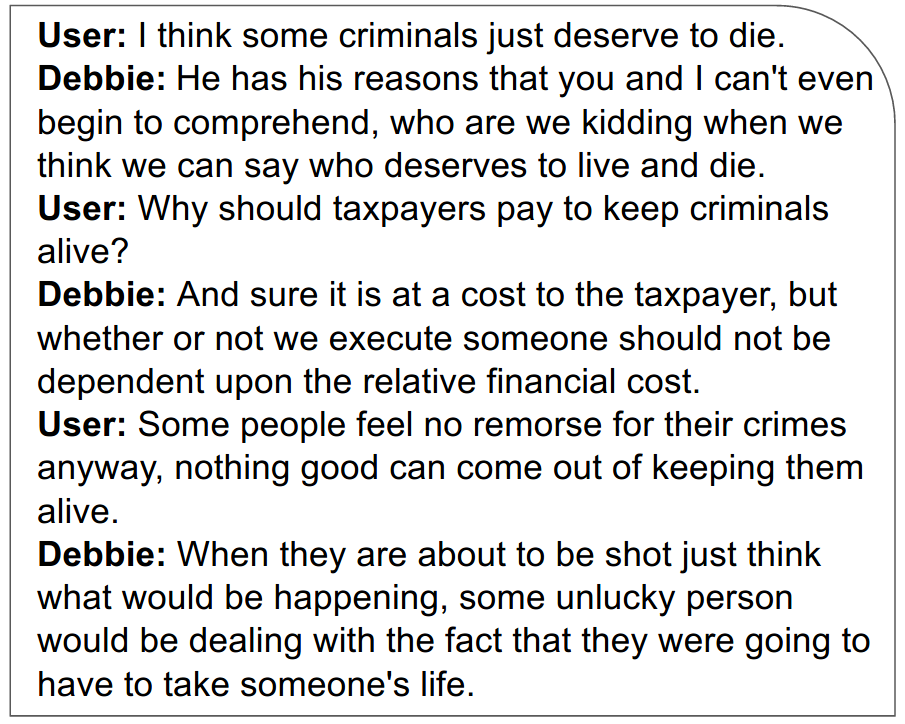}
\end{center}
\vspace{-10pt}
\caption{Chat where Debbie is against the death penalty}
\vspace{-15pt}
\label{fig:debbie_output_dp}
\end{wrapfigure}

We further optimized our algorithm by implementing a graph-based comparison method to find an acceptable cluster faster. We create a graph with the cluster heads as nodes. We start at a random head and find how similar it is to the input. If the similarity passes a high threshold of 0.9, we use the related cluster automatically. Otherwise, if the similarity is very high, say, above a threshold of 0.8, we eliminate connected edges where the similarity is very low, below a threshold of 0.5. Conversely, if the similarity is very low, say, below 0.5, we eliminate connected edges where the similarity is very high, above 0.8. In the case where we don't find a head which surpasses our high threshold, we continue to explore our graph until all the clusters have either been visited or eliminated from consideration. We then select the head with the highest similarity score.

\section{Evaluation}
\label{sec:evaluation}

\begin{wraptable}{r}{0.5\textwidth}
\vspace{-20pt}
\begin{tabular}{|c|c|c|c|c|}
\hline
{\bf topic} & {\bf stance} & \begin{tabular}[c]{@{}c@{}}{\bf baseline} \end{tabular} & \begin{tabular}[c]{@{}c@{}}{\bf cluster} \end{tabular} & \begin{tabular}[c]{@{}c@{}} {\bf graph}  \end{tabular} \\ \hline
\begin{tabular}[c]{@{}c@{}}DP\end{tabular} & for & 60.6 & 3.9 & 8.1 \\ \hline
\begin{tabular}[c]{@{}c@{}}DP\end{tabular} & against & 55.5 & 7.9 & 5.0 \\ \hline
\begin{tabular}[c]{@{}c@{}}GC\end{tabular} & for & 70.7 & 25.2 & 24.3 \\ \hline
\begin{tabular}[c]{@{}c@{}}GC\end{tabular} & against & 73.5 & 22.9 & 15.3 \\ \hline
\begin{tabular}[c]{@{}c@{}}GM\end{tabular} & for & 62.8 & 10.0 & 9.2 \\ \hline
\begin{tabular}[c]{@{}c@{}}GM\end{tabular} & against & 62.8 & 3.2 & 2.9 \\ \hline
\end{tabular}
\caption{Average response times in seconds}
\vspace{-20pt}
\label{table:average_times}
\end{wraptable}

A sample conversation with Debbie is shown in Fig \ref{fig:debbie_output_dp}, where the user supports the death penalty and Debbie opposes it. For a start, we looked at methods for faster response retrieval and the quality of the responses. The most basic (baseline) method just finds the similarity score between the input and each possible response in our database, and returns the response with the highest similarity score. The second method is the simple clustering method and the third method is clustering with the graph method described in Section \ref{sec:methodology}.

Table \ref{table:average_times} shows the average response times for each retrieval method. 
We arrived at these by testing for three sentences per stance per topic, each deliberately chosen such that they would access different clusters. As expected, both the cluster and the cluster graph methods perform faster than the baseline. The cluster graph method is faster than using just clusters in most cases. The exception was the "for" case of death penalty, which, we believe, can be attributed to the significantly larger size of the cluster that is accessed by the cluster graph, triggering a greater number of computations. Given that our database only has high quality arguments, the appropriateness of Debbie's responses are primarily dependent on the performance of the similarity algorithm. However, exchange of only high quality arguments between the system and the user, with minimal repetition (none from Debbie) hampers the natural flow of the conversation.


\section{Future Work}
\label{sec:future}

The prototype we are proposing represents our pilot work with Debbie, an argumentative chatbot. Our work in this domain is still in progress and we have a lot of future work planned based on our preliminary observations. We acknowledge the fact that we must migrate Debbie away from the assumption that the argument as a whole will consist of argumentatively sound statements. In our evaluation, we observed a high usage of utterances such as \textit{You're just wrong.} and \textit{I don't think so.} from the user. These statements have low argumentative quality, or, are completely off-topic. Hence, responding to them with a high quality argument tended to sound unnatural and inappropriate. Therefore, in order to make the conversation sound less robotic and more natural, we must detect user utterances which are not argumentatively sound and respond accordingly. 

Lukin et al.(2017) talk about the role of personality in persuasion \cite{lukinargument} and Bowden et al.(2016) have shown that adding a layer of stylistic variation to a dialogue is sufficient for representing personality between speakers \cite{Bowden2016M2DMT}. We intend to investigate how we can enhance the user's experience by entraining Debbie's personality with respect to the user's personality. While our initial results are promising, we need to improve Debbie's performance with regards to retrieval time. We can potentially do this by recursively employing the graph method within the clusters - similar to hierarchical clusters. We also intend to explore alternative information retrieval methods such as indexing, to create a more balanced trade-off between appropriateness and response time. Debbie currently uses the UMBC STS for calculating similarity scores, which is known to not work very well for argumentative sentences \cite{misra:2016}. We intend to use a better argument similarity algorithm in the future. \\

\bibliographystyle{plain}
\bibliography{author}

\end{document}